\documentclass[nohyperref]{article}

\usepackage{microtype}

\usepackage{graphicx}
\graphicspath{{./figs/}}
\usepackage{subcaption} 
\usepackage{booktabs} 

\usepackage{hyperref}

\usepackage[accepted]{icml2022}

\usepackage{amsmath}
\usepackage{amssymb}
\usepackage{mathtools}
\usepackage{amsthm}
\usepackage{color}

\usepackage[capitalize,noabbrev]{cleveref}

\theoremstyle{plain}

\theoremstyle{definition}

\theoremstyle{remark}

\usepackage[textsize=tiny]{todonotes}

\usepackage{tikz}
\newcommand{\cblock}[3]{
  \hspace{-1.5mm}
  \begin{tikzpicture}
    [
    node/.style={square, minimum size=10mm, thick, line width=0pt},
    ]
    \node[fill={rgb,255:red,#1;green,#2;blue,#3}] () [] {};
  \end{tikzpicture}%
}

\newcommand{\xorls}{\textsc{E}x\textsc{ORL}}

\icmltitlerunning{Exploratory Data for Offline Reinforcement Learning}

\begin{document}

\twocolumn[
\icmltitle{Don't Change the Algorithm, Change the Data: \\
Exploratory Data for Offline Reinforcement Learning}
\vspace{-0.3cm}





\icmlsetsymbol{equal}{*}

\begin{icmlauthorlist}
\icmlauthor{Denis Yarats}{equal,nyu,fair}
\icmlauthor{David Brandfonbrener}{equal,nyu}
\icmlauthor{Hao Liu}{ucb}
\icmlauthor{Michael Laskin}{ucb}\\
\icmlauthor{Pieter Abbeel}{ucb}
\icmlauthor{Alessandro Lazaric}{fair}
\icmlauthor{Lerrel Pinto}{nyu}
\end{icmlauthorlist}

\icmlaffiliation{nyu}{Department of Computer Science, New York University}
\icmlaffiliation{fair}{Facebook AI Research}
\icmlaffiliation{ucb}{Department of Electrical Engineering and Computer Science, UC Berkeley}

\icmlcorrespondingauthor{Denis Yarats}{denisyarats@nyu.edu}
\icmlcorrespondingauthor{David Brandfonbrener}{david.brandfonbrener@nyu.edu}

\icmlkeywords{Offline RL, Batch RL, Exploration, Unsupervised Exploration, Data, Diversity}

\vskip 0.3in
]



\printAffiliationsAndNotice{\icmlEqualContribution} 

\newif\ifincludeappendix

\begin{abstract}
    Recent progress in deep learning has relied on access to large and diverse datasets. Such data-driven progress has been less evident in offline reinforcement learning (RL), because offline RL data is usually collected to optimize specific target tasks limiting the data’s diversity. In this work, we propose Exploratory data for Offline RL (\xorls), a data-centric approach to offline RL.~\xorls~first generates data with unsupervised reward-free exploration, then relabels this data with a downstream reward before training a policy with offline RL. We find that exploratory data allows vanilla off-policy RL algorithms, without any offline-specific modifications, to outperform or match state-of-the-art offline RL algorithms on downstream tasks. Our findings suggest that data generation is as important as algorithmic advances for offline RL and hence requires careful consideration from the community. Code and data can be found at \url{https://github.com/denisyarats/exorl}.
\end{abstract}

\vspace{-0.5cm}
\section{Introduction}

Large and diverse datasets have been the cornerstones of many impressive successes at the frontier of machine learning, such as image recognition~\citep{ deng2009imagenet, krizhevsky2012imagenet, radford2021clip}, natural language processing~\citep{radford2019language, raffel2019exploring, wang2019glue, brown2020language}, and protein folding~\citep{uniprot2020,jumper2021highly}.
Training on diverse datasets often dramatically improves performance and enables impressive generalization capabilities.


In reinforcement learning (RL), however, the potential of large and diverse datasets has not yet been fully realized. The historically dominant approach in RL has been to train online from scratch by interacting with a task-specific environment~\citep{sutton2018reinforcement, mnih2013playing}. More recently there has been increased interest in learning offline from fixed datasets \citep{levine2020offline,fujimoto2019off-policy}. 
This line of offline RL research has been accelerated by benchmark datasets such as D4RL \citep{fu2020d4rl} and RL Unplugged \citep{gulcehre2020rl}.

\begin{figure}[t!]
    \vspace{-0.5cm}
    \centering
    \begin{subfigure}{\linewidth}
        \centering
        \includegraphics[width=\textwidth]{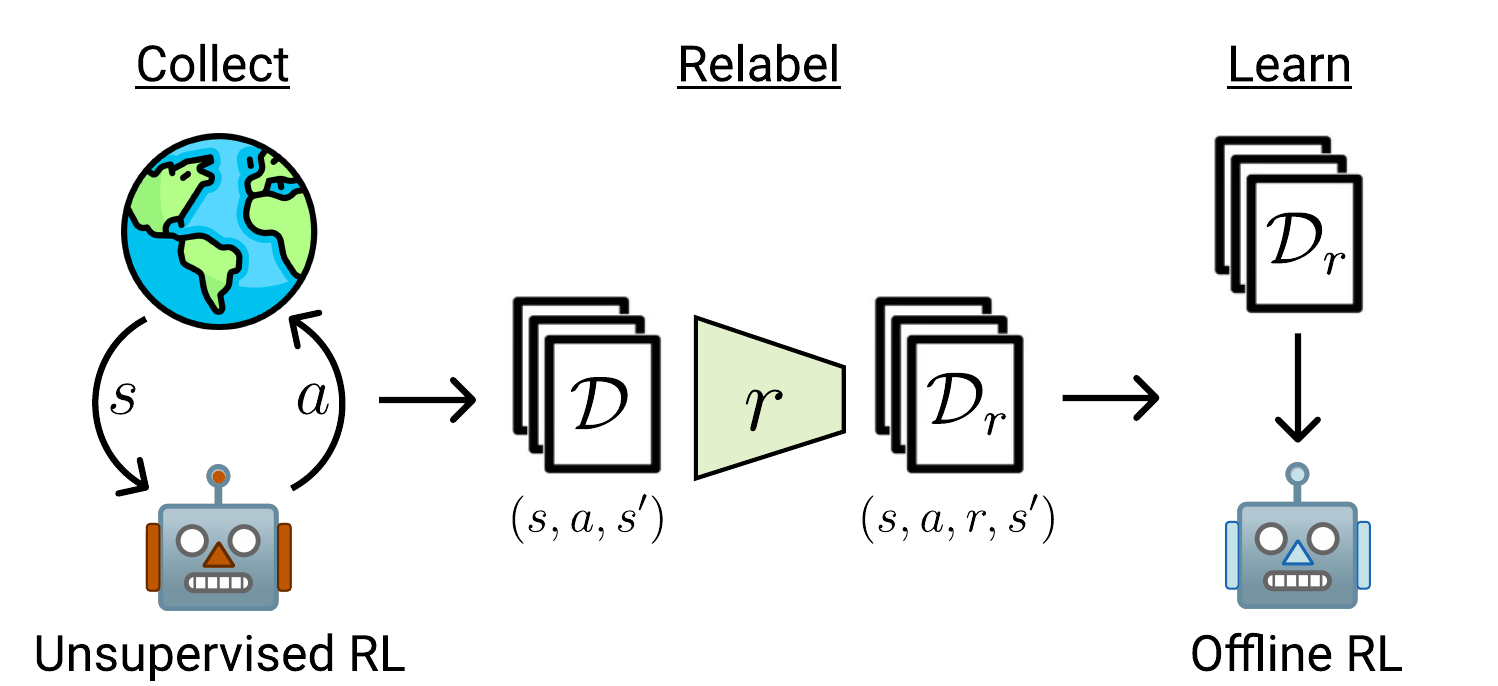}
        \caption{The three phases of~\xorls: data collection with unsupervised RL, data relabeling with the downstream reward, and learning on the labeled data with offline RL.}
        \label{fig:cartoon}
    \end{subfigure}
    \begin{subfigure}{\linewidth}
        \centering
        \includegraphics[width=0.8\linewidth]{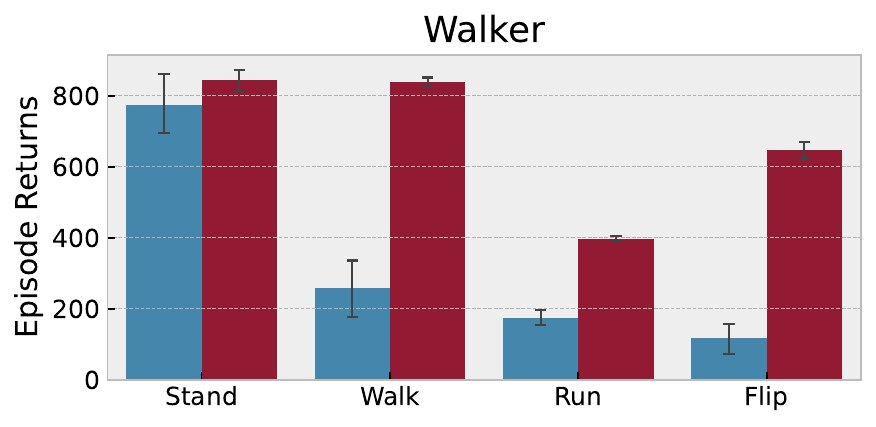}\\
        \hspace{1.5mm} \cblock{52}{138}{189}\hspace{1mm}Supervised\hspace{1.5mm}
    \cblock{166}{6}{40}\hspace{1mm}Unsupervised\hspace{1.5mm}
    
    \caption{Offline RL evaluation using vanilla TD3 of supervised data (the replay buffer of a TD3 agent trained on Walker Stand) versus unsupervised exploratory data (collected reward-free on Walker with Proto). Unsupervised data demonstrates an average performance improvement of $106\%$ across the four evaluation tasks. 
    }
    \label{fig:unsup_vs_sup}  
    \end{subfigure}
    \caption{Summary of the \xorls~framework.}
    \vspace{-0.8cm}
\end{figure}

\begin{figure*}[h]
   \centering
	\begin{subfigure}{\linewidth}
		\centering
		\includegraphics[width=0.8\linewidth]{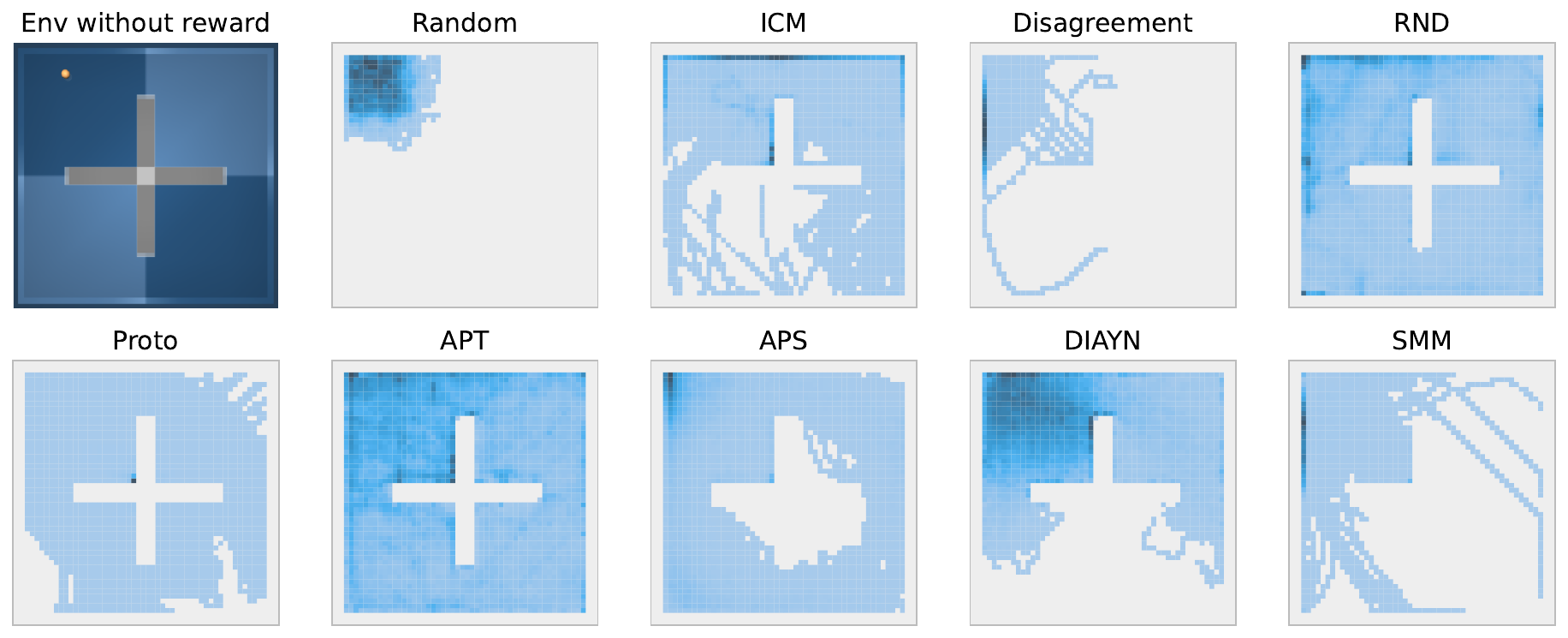}
		\caption{\textbf{Collect}: an unsupervised RL agent interacts with the environment in a reward-free manner and stores the acquired transitions into a dataset $ \mathcal{D}$ of $ (s,a,s')$ tuples. We visualize state distributions of the pointmass for each collected dataset $\mathcal{D}$.}
        \label{fig:pm_data}
  
   \end{subfigure}
    \centering

	\begin{subfigure}{\linewidth}
		\centering
		\includegraphics[width=0.9\linewidth]{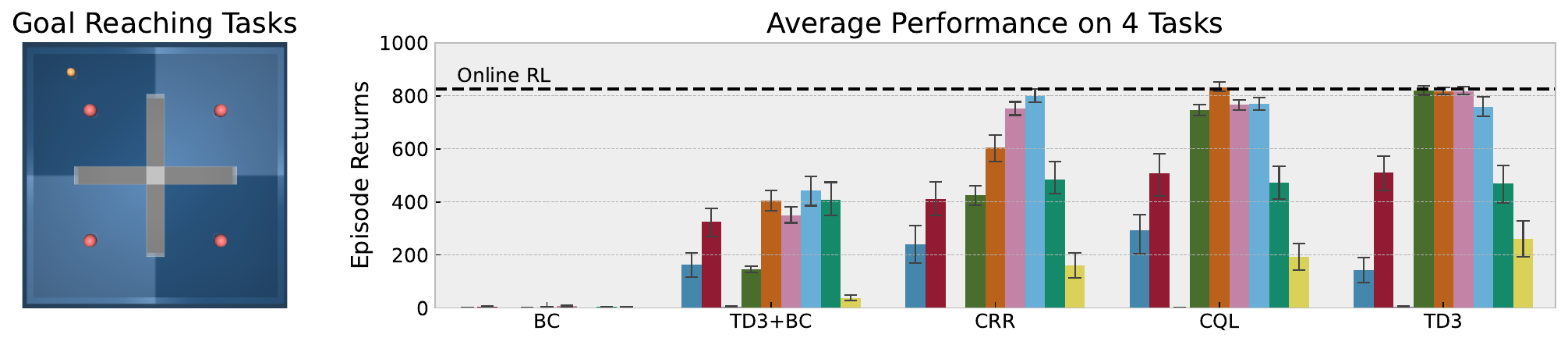}\\
		\hspace{1.5mm}
    \cblock{52}{138}{189}\hspace{1mm}Random\hspace{1.5mm}
    \cblock{166}{6}{40}\hspace{1mm}ICM\hspace{1.5mm}
    \cblock{122}{104}{166}\hspace{1mm}Disagreement\hspace{1.5mm}
    \cblock{70}{120}{33}\hspace{1mm}RND\hspace{1.5mm}
    \cblock{213}{94}{0}\hspace{1mm}Proto\hspace{1.5mm}
    \cblock{204}{121}{167}\hspace{1mm}APT\hspace{1.5mm}
    \cblock{86}{180}{233}\hspace{1mm}APS\hspace{1.5mm}
    \cblock{0}{158}{115}\hspace{1mm}DIAYN\hspace{1.5mm}
    \cblock{240}{228}{66}\hspace{1mm}SMM

		\caption{\textbf{Relabel} and \textbf{Learn}: Each tuple in $ \mathcal{D} $ is relabeled by a task-specific reward function to yield a new dataset $ \mathcal{D}_r$ of $ (s,a,r,s')$ tuples (there are $4$ different rewards in this example, one for each goal). Finally, we run an offline RL algorithm on each dataset $ \mathcal{D}_r$ in order to learn a policy to maximize the specified reward. Here, offline RL is performed by various state of the art offline RL algorithms as well as standard supervised learning (BC) and off-policy RL (TD3).}
        \label{fig:pm_eval}
        
   \end{subfigure}

    \caption{Our \xorls~framework exploratory data collection for offline RL consists of three phases:  \textit{collect}  \textbf{(a)}, \textit{relabel} and \textit{learn} \textbf{(b)}. Here we showcase our method on a pointmass maze environment to demonstrate that it is possible to collect exploratory data in a completely unsupervised manner that enables effective multi-task offline RL.}
    \label{fig:pm}
\end{figure*}

Currently, offline RL datasets are collected by training task-specific RL algorithms and saving the data visited during the training process. Offline RL algorithms are then evaluated on the same task that the dataset was created with. 
As a result, very strong performance on these benchmarks can often be achieved with simple methods like behavior cloning~\cite{Pomerleau1988ALVINNAA, chen2021dt} or one step of policy iteration \cite{brandfonbrener2021offline, gulcehre2021regularized} without requiring substantial amounts of planning.
This raises a fundamental question for offline RL -- What happens when your data does not come from the same task?


In this work, we propose a new framework for offline RL that can test the ability of offline RL algorithms to generalize to tasks that are not observed during the creation of the dataset. This framework, which we call Exploratory data for Offline RL (\xorls), has
three steps: collect, relabel, and learn.
To instantiate \xorls, we \emph{collect} datasets using nine different unsupervised exploration strategies~\citep{laskin2021urlb} on several environments from the DeepMind control suite \cite{tassa2018deepmind}. Then, we \emph{relabel} the data with different reward functions. Finally, we \emph{learn} a policy by running several different offline RL algorithms for each relabeled dataset.
A schematic of the framework can be found in \cref{fig:cartoon} and an instantiation in a simple pointmass maze environment in~\cref{fig:pm}.

We provide a thorough analysis of current offline RL methods by running large-scale experiments with \xorls. The main findings of these experiments are:
\begin{enumerate}
    \item With sufficiently diverse exploratory data, vanilla off-policy RL agents can effectively learn offline, and even outperform carefully designed offline RL algorithms. This suggests that advances in data collection are as important as algorithmic advances for offline RL. 
    \item Compared to data collected in a supervised manner (i.e. for a task-specific reward function), the unsupervised exploratory data allows offline RL algorithms to learn multiple behaviors from a single \xorls~dataset. For example, unsupervised exploration yields a 106\% performance increase over supervised data on multi-task learning in the Walker environment (\cref{fig:unsup_vs_sup}).
\end{enumerate}

Finally, \xorls~will be released as a benchmark suite of exploratory datasets to test offline RL algorithms in a substantially different regime than prior benchmark suites. While prior work focuses on datasets with limited coverage, our unsupervised datasets are collected to maximize various notions of coverage or diversity. 
We should emphasize that we are not saying that these new datasets are better offline RL benchmarks, just that they provide a significantly different and important challenge on the road to useful offline RL.

\section{Related Work}

\subsection{Datasets for offline RL}

The most closely related line of work is that of data-collection and benchmarking in offline RL. Both the D4RL \citep{fu2020d4rl} and RL Unplugged \citep{gulcehre2020rl, mathieu2021starcraft} benchmark suites consist of datasets collected by policies attempting to optimize task-specific rewards. These datasets are either replay buffers of the training run or trajectories collected by a snapshot from somewhere in training. 
D4RL also includes some demonstration and goal-directed data. Importantly, the datasets are then tested on the same reward they were trained on (or the demonstrations they were targeting). 

In contrast, our contribution is to use unsupervised exploratoy data collection to test offline RL algorithms. This creates datasets that facilitate multi-task offline RL from a single dataset and poses a very different challenge for offline RL than prior benchmarking work. 

\subsection{Unsupervised pre-training for RL}

Recently, URLB \citep{laskin2021urlb} proposed a benchmarking protocol for unsupervised pre-training algorithms in RL. In that work, each pre-training algorithm is allowed a fixed budget of unsupervised interactions with the environment and then oututs a pre-trained policy that is used to initialize an online RL algorithm at test time. 
In our work, instead of using a pre-trained policy as the output of the pre-training phase and then training online, we use the \emph{dataset} as the output of the pre-training phase and then train entirely \emph{offline}. 
To perform this data collection, we consider the same eight pre-training algorithms as in URLB \citep{pathak2017curiosity, burda2018exploration, eysenbach2018diversity, pathak2019self, lee2019efficient, liu2021behavior, liu2021aps, yarats2021reinforcement} as well as a uniformly random data collector.

There is also a line of purely theoretical work that studies the ``reward-free RL'' problem in the tabular and linear cases \cite{jin2020reward, wang2020reward}. Much like the goal of our protocol, these papers focus on collecting a dataset that allows for offline learning of a near optimal policy for any reward function. Our work can be viewed as an empirical evaluation of several different ways to scale up these reward-free RL algorithms to large domains with neural networks.

\subsection{Multi-task reinforcement learning}

Another important related line of work focuses on sharing data across tasks. Here, data collection is generally directed at multiple different goals with some combination of demonstrations and standard RL with task-specific rewards \cite{riedmiller2018learning, cabi2019scaling, kalashnikov2021mt, yu2021conservative, chen2021batch}. 
In contrast, we focus on studying the capabilities of purely unsupervised data collection. We view these two research directions as complementary. For example, they could potentially be combined in future work, where unsupervised data could be added to multi-task data to facilitate transfer. 

Another line of work generates goal-based multi-task data using different goals as the different tasks where the goals can be generated in an unsupervised manner \citep{andrychowicz2017hindsight, ecoffet2019go, dendorfer2020goal, mendonca2021discovering}. \citet{endrawis2021efficient} in particular uses goal-based task generation to collect exploratory data to use with offline RL on robotic tasks.
In contrast, we do not use any goal-based and focus our efforts on understanding how offline RL methods perform in the exploratory setting.

We should also note that our relabeling procedure is not novel and similar ideas have been used widely in the multi-task and goal-based RL literature \citep{andrychowicz2017hindsight, riedmiller2018learning, sekar2020planning, eysenbach2020rewriting, li2020generalized, kalashnikov2021mt}.

\subsection{Decoupling exploration and exploitation}

A few recent position papers argue for separating the exploration and exploitation problems to facilitate scalable reinforcement learning \citep{levine2021understanding, riedmiller2021collect}. Ideally, collecting large datasets in an unsupervised manner and re-using the data across many downstream tasks can dramatically improve data-efficiency on a per-task basis. Moreover, by splitting the problem into two, we can more effectively isolate the effects of each component of the algorithm. Our paper can be seen as taking a first step towards instantiating this vision of effectively decoupling exploration and exploitation.

\subsection{Concurrent work}
Concurrently to our paper, \cite{lambert2022challenges} proposes a similar framework for unsupervised exploration followed by offline RL. In contrast to our work, they focus on proposing a novel MPC-based exploration technique and do not compare multiple offline RL algorithms in the exploratory setting, instead they only use CRR. 



\section{Our framework: \xorls}

In this section we explicitly define our framework for exploratory data collection for offline RL (\xorls). The framework consists of three steps: collect, relabel, and learn. The rest of this section goes through each step in detail and the protocol is summarized in~\cref{alg}.

\begin{algorithm}[h]
   \caption{ExORL: Collect-Relabel-Learn}
   \label{alg}
\begin{algorithmic}
   \STATE {\bfseries Input:} trajectory budget $ B$, environment $ \mathcal{E} $, data collection algo $\mathcal{C}$, reward function $ r$, offline RL algo $ \mathcal{O}$  
   \STATE {\bfseries Collect:} Run $ \mathcal{C}$ on $ \mathcal{E}$ to collect $ B$ trajectories into a dataset $ \mathcal{D}$ of $ (s,a,s')$ tuples.
   \STATE {\bfseries Relabel:} Update each tuple in $ \mathcal{D} $ by using $ r $ to get $ (s,a,r(s,a), s') $ tuples. Call the relabeled dataset $ \mathcal{D}_r$.
   \STATE {\bfseries Learn:} Run $\mathcal{O}$ on the labeled data in $ \mathcal{D}_r$.
   
\end{algorithmic}
\end{algorithm}

We assume that we have access to a reward-free episodic MDP environment $\mathcal{E}$ with state space $ \mathcal{S}$, action space $ \mathcal{A}$, and stochastic transition dynamics determined by $ P(\cdot|s,a)$. In our problem setup we are given an interaction budget $ B $ that limits the number of trajectories that we can collect in the collection phase. Rewards for relabeling can be defined by any function $ r: \mathcal{S} \times \mathcal{A} \to \mathbb{R}$.

\paragraph{Collect.} The first phase of \xorls~is to collect the unlabeled exploratory data. To do this, we can use any algorithm $ \mathcal{C}$ that at each episode $ k $ outputs a policy $ \pi_k $ conditioned on the history of prior interactions. Each episode collected by $ \mathcal{C}$ is stored in the dataset $ \mathcal{D}$ as a series of $ (s,a,s')$ tuples. We run the collection algorithm $ \mathcal{C}$ until we have collected $ B$ episodes into the dataset $ \mathcal{D}$.


In practice, we evaluate nine different unsupervised data collection algorithms as $ \mathcal{C}$ in this phase. The first is a simple baseline that always outputs the uniformly random policy. The remaining eight are taken from URLB \citep{laskin2021urlb} and can be sorted into three groups: 
\begin{itemize}
    \item Knowledge-based methods that maximize error of a predictive model: ICM \citep{pathak2017curiosity}, Disagreement \citep{pathak2019self}, RND \citep{burda2018exploration}.
    \item Data-based algorithms that maximize some estimate of coverage of the state space: APT \citep{liu2021behavior} and ProtoRL \cite{yarats2021reinforcement}.
    \item Competence-based algorithms that learn a diverse set of skills: DIAYN \cite{eysenbach2018diversity}, SMM \cite{lee2019efficient}, and APS \citep{liu2021aps}.
\end{itemize}

\paragraph{Relabel.} Having collected a dataset $ \mathcal{D}$ of $ (s,a,s') $ tuples, the next phase of our protocol relabels the data using the given reward function $ r$. This simply requires evaluating $ r(s,a) $ at each tuple in the dataset and then adding $ (s,a,r(s,a), s')$ to the relabeled dataset $ \mathcal {D}_r$.

In practice, we use either standard or hand-designed reward functions in each environment~\citep{tassa2018deepmind, laskin2021urlb}, and are detailed in ~\cref{app:dmc_tasks}. We note, however, that our framework also allows for learning reward function (i.e. inverse RL or successor features).

\begin{figure*}[t]
    \centering

		\includegraphics[width=\linewidth]{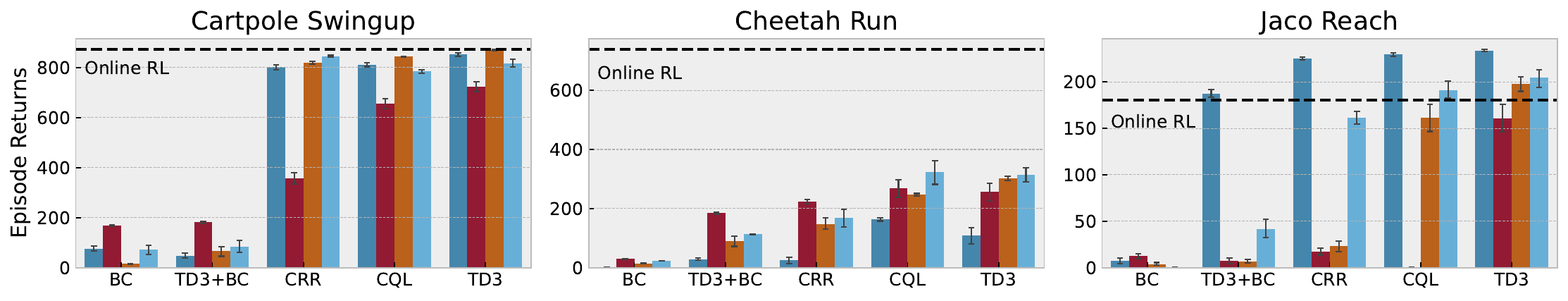}
		 \hspace{1.5mm}
    \cblock{52}{138}{189}\hspace{1mm}Random\hspace{1.5mm}
    \cblock{166}{6}{40}\hspace{1mm}ICM\hspace{1.5mm}
    \cblock{213}{94}{0}\hspace{1mm}Proto\hspace{1.5mm}
    \cblock{86}{180}{233}\hspace{1mm}APS\hspace{1.5mm}
		\caption{Offline evaluation of unsupervised datasets on one task for each of three different domains. Here we choose four representative unsupervised exploration algorithms, for full results see \cref{section:mult_env_full}. Vanilla TD3 usually outperforms all three offline RL algorithms. }
        \label{fig:mult_env}
  
\end{figure*}

\paragraph{Learn.} Finally, we can learn a policy by running an offline RL algorithm $ \mathcal{O}$ on our labeled dataset $ \mathcal{D}_r$. The offline RL algorithm learns entirely offline by sampling tuples from the dataset and then outputs the final policy to be evaluated online in the environment $ \mathcal{E}$ by calculating the return with respect to the reward function $ r$.  

In practice, we run five different offline RL algorithms for each reward function on each dataset. As a baseline we run simple behavior cloning. Then we run three different state of the art offline RL algorithms that each use a different mechanism to prevent extrapolation beyond the actions in the data. These algorithms are: CRR \citep{wang2020critic} which uses filtered behavior cloning, CQL \citep{kumar2020conservative} which uses pessimistic Q estimates, and TD3+BC \citep{fujimoto2021minimalist} which regularizes toward the behavior. Finally, we run standard TD3 \citep{fujimoto2018addressing} as a baseline to test what happens when we run an off-policy RL algorithm that was originally designed for the online setting and has no mechanism explicitly designed to prevent extrapolation in the offline setting.

\begin{figure*}[t]
    \centering

		\includegraphics[width=\linewidth]{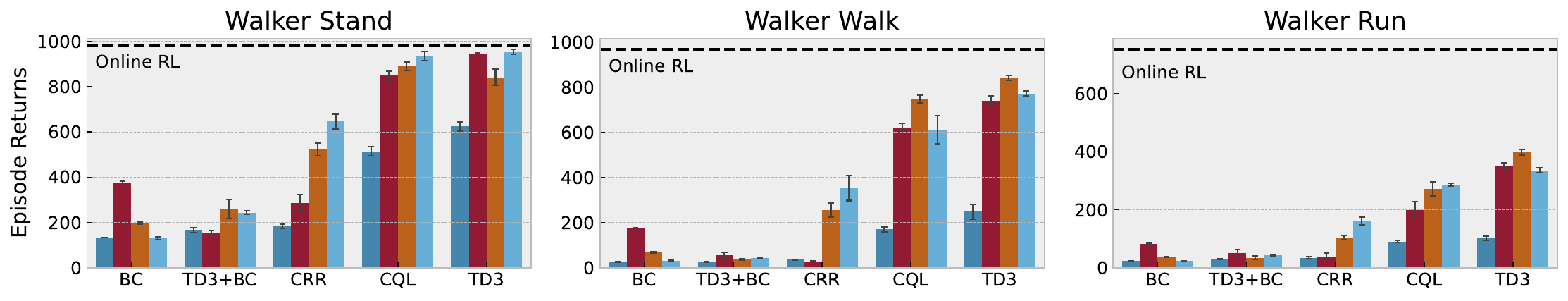}

   \hspace{1.5mm}
    \cblock{52}{138}{189}\hspace{1mm}Random\hspace{1.5mm}
    \cblock{166}{6}{40}\hspace{1mm}ICM\hspace{1.5mm}
    \cblock{213}{94}{0}\hspace{1mm}Proto\hspace{1.5mm}
    \cblock{86}{180}{233}\hspace{1mm}APS\hspace{1.5mm}

    	\caption{Offline evaluation of datasets for the Walker environment under three different rewards (Stand, Walk, and Run). We observe that \xorls~allows for data relabeling to enable multi-task offline RL. See~\cref{section:mult_task_full} for full results across all exploration algorithms.}
        \label{fig:mult_task}

    \label{fig:main}
\end{figure*}

\section{Experiments}

In this section we conduct an empirical study to answer the following questions:
\textbf{Q1}: Can the diversity of unsupervised data collected by~\xorls~enable vanilla off-policy RL agents to work in the offline setting? \textbf{Q2}: Is it possible to relabel this unsupervised data to facilitate multi-task offline RL? \textbf{Q3}: Is exploratory data necessary for multi-task offline RL? 
\textbf{Q4}: Would it be useful to mix exploratory data with task-specific data? \textbf{Q5}: How effective is~\xorls~as we scale the data collection budget?

\subsection{Setup}
We evaluate our framework on a set of challenging environments from the DeepMind control suite~\citep{tassa2018deepmind}. Each environment has several different rewards that we use to evaluate multi-task generalization. See~\cref{app:dmc_tasks} for detailed overview of the environments. We use URLB~\citep{laskin2021urlb} for unsupervised data collection with a budget of 1M reward-free environment interactions ($1000$ episodes) unless stated otherwise. The transitions in these datasets are then relabeled under a specific reward and used for offline RL. For offline RL we use five algorithms: BC, TD3+BC, CRR, CQL, and TD3. We adhere closely to the original hyper-parameter settings for each algorithm, but in several cases we perform hyper-parameter tuning to achieve best possible performance. See~\cref{app:hyperparams} for more details. We train offline RL algorithms for 500k gradient updates and then evaluate by rolling out $10$ episodes in the environment. We report mean and standard error across $10$ random seeds. A description of computational requirements is provided in~\cref{app:compute}.

\subsection{Comparing offline RL algorithms on \xorls}

\textbf{Q1}: Can the diversity unsupervised data collected by~\xorls~enable vanilla off-policy RL agents to work in the offline setting?
\textbf{A1}: \textit{Yes, vanilla off-policy RL agents can perform well from \xorls~data, in particular vanilla TD3 outperforms the other offline RL algorithms.}. 

To verify this, we took three environments (Cartpole, Cheetah, and Jaco) and one reward for each environment (Swingup, Run, and Reach respectively). For each environment we run all nine exploration algorithm, but for clarity only present four in the main text that represent each of the main families of algrithms (ICM: knowledge-based, Proto: data-based, APS: competence-based, and Random as a baseline). We then evaluate with five different offline RL algorithms. Results are in~\cref{fig:mult_env} and~\cref{section:mult_env_full}.

Specifically, comparing the results on these datasets we see that vanilla TD3 consistently outperforms more sophisticated offline RL algorithms (TD3+BC, CRR, and CQL). This is a somewhat surprising result because the offline RL algorithms were designed specifically for the offline setting while TD3 was not. This emphasizes how the unsupervised data collected by \xorls~provides a substantially different setting than the traditional offline RL problems that the algorithms were designed for. We hypothesize that this difference occurs because \xorls~produces diverse datasets that allow for more aggressive trajectory stitching and less severe extrapolation than prior datasets.

While the relative strength of TD3 is consistent, the results are somewhat more varied across the different data collection algorithms. For example, in Jaco, the random data collection surprisingly outperforms all of the unsupervised exploration algorithms. This is due to over-exploration by the unsupervised algorithms which causes the arm to explore low-reward positions outside of the workspace, while the random agent keeps the hand closer to initialization.
Also, while \xorls~achieves performance comparable to or exceeding online RL in Cartpole Swingup and Jaco Reach, the performance on Cheetah Run is still not competitive with the optimal policy. This is likely due to insufficient exploration of the higher-dimensional state space in regions of high reward, suggesting there is still room to improve unsupervised exploration algorithms.


\begin{figure*}[h]
    \centering

    \includegraphics[width=0.9\linewidth]{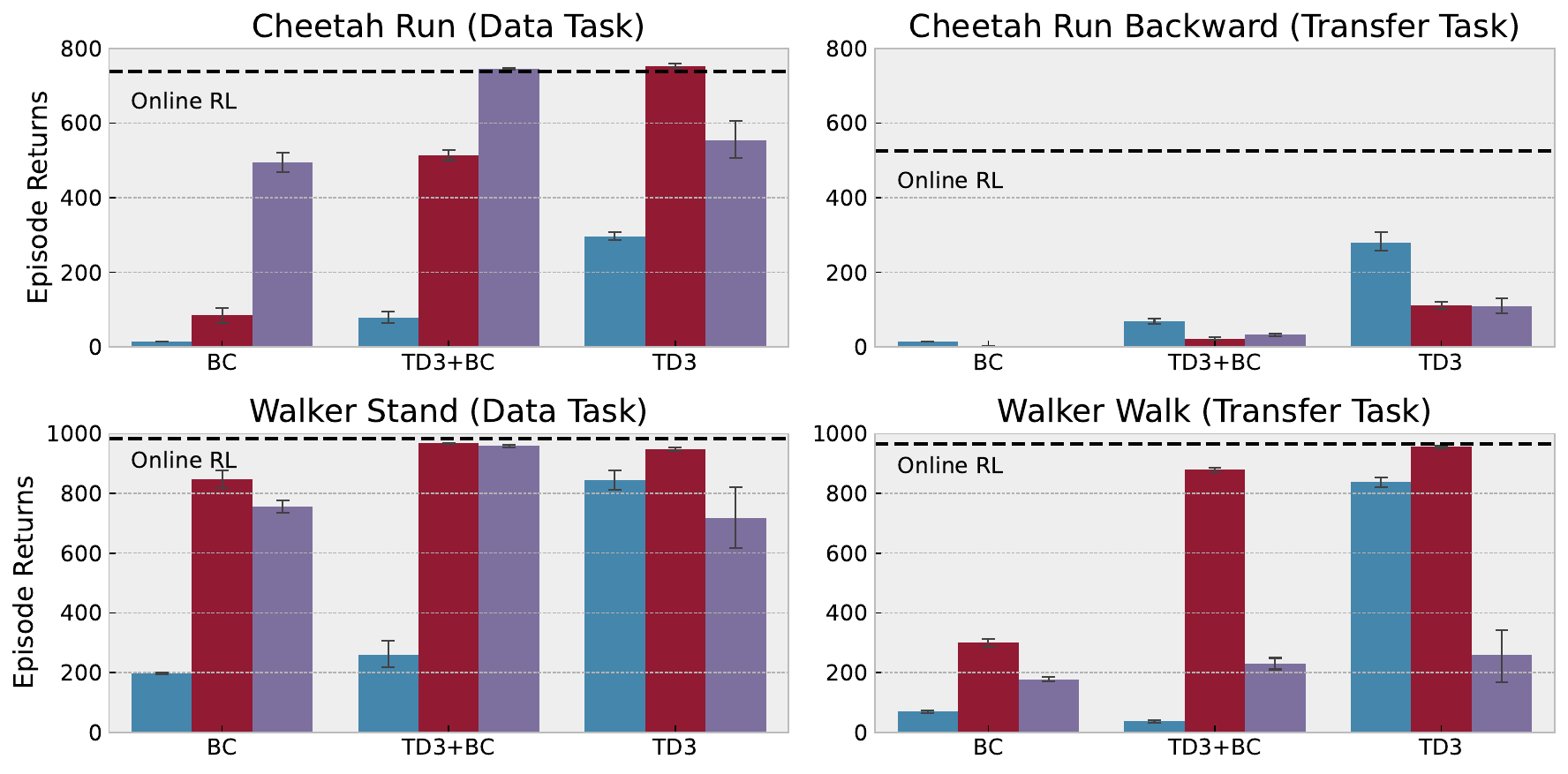}\\

   \hspace{1.5mm}
    \cblock{52}{138}{189}\hspace{1mm}Unsupervised\hspace{1.5mm}
    \cblock{166}{6}{40}\hspace{1mm}Semi-supervised\hspace{1.5mm}
    \cblock{122}{104}{166}\hspace{1mm}Supervised\hspace{1.5mm}

    \caption{A comparison of three different data-collection strategies: unsupervised (intrinsic reward only), semi-supervised (intrinsic reward plus data task reward), and supervised (data task reward only). Data diversity is key to enable more aggressive trajectory stitching and task transfer. }
    \label{fig:supervision}
\end{figure*}

\subsection{Multi-task learning with \xorls}

\textbf{Q2}: Is it possible to relabel this unsupervised data to facilitate multi-task offline RL? \textbf{A2:} \textit{Yes, multi-task offline RL can perform well from \xorls~data}.

To confirm this we evaluate on one environment (Walker) across three different tasks (Stand, Walk, and Run). This means that each data collection algorithm only collects one dataset and we relabel it with three different rewards. Results are in~\cref{fig:mult_task}.

In particular, we are able to nearly solve both the Stand and Walk tasks from a single dataset collected with no knowledge of the rewards. Performance on Run is further away from optimal, but still strong enough to suggest meaningful progress towards the desired behavior. So, while there is room for improvement of the unsupervised exploration, the \xorls~framework has clear promise to learn multiple tasks from a single task-agnostic dataset.

Again, we observe consistently strong performance of vanilla TD3 on exploratory data, confirming the result from the prior experiment. 
And comparing the different exploration algorithms, we see similar performance across each of the three representative algorithms (ICM, Proto, and APS) with each of them being much stronger than random.

\begin{figure*}[h]
    \centering

    \includegraphics[width=0.9\linewidth]{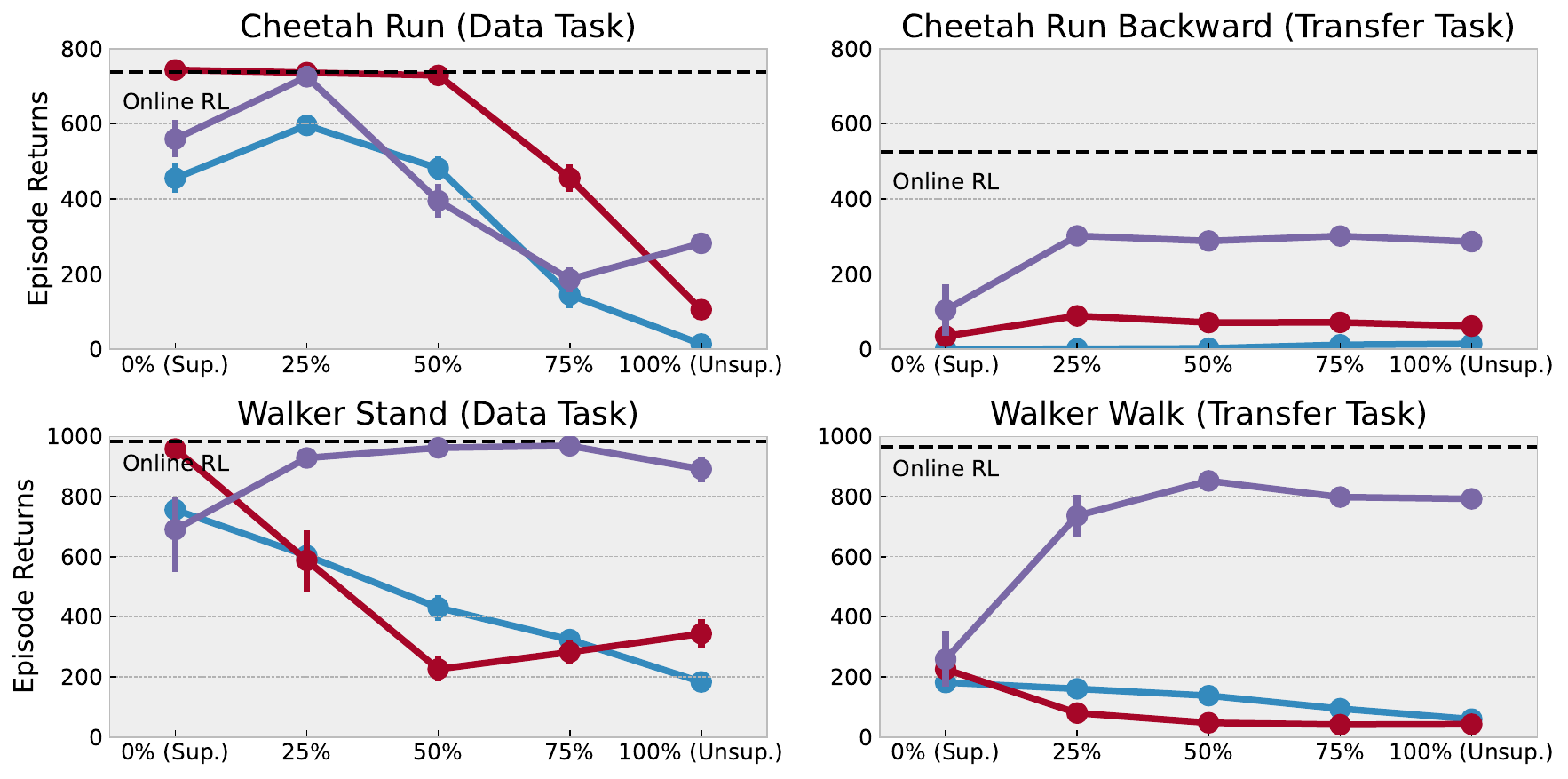}\\
        \hspace{1.5mm} \cblock{52}{138}{189}\hspace{1mm}BC\hspace{1.5mm}
    \cblock{166}{6}{40}\hspace{1mm}TD3+BC\hspace{1.5mm}
    \cblock{122}{104}{166}\hspace{1mm}TD3\hspace{1.5mm}
    
    \caption{The effect of mixing supervised data (collected by TD3 on the data tasks) with unsupervised data (collected reward-free by Proto). Unsupervised data allows easier transfer and more efficient generalization to unseen tasks.}
    \label{fig:mixing}
     \vspace{-0.5cm}
\end{figure*}

\subsection{Comparing \xorls~to supervised data collection}

\textbf{Q3}: Is exploratory data necessary for multi-task offline RL? 
\textbf{A3:} \textit{Yes, the unsupervised exploratory data facilitates transfer where the supervised data does not}.

To demonstrate this, we consider different data-collection strategies ranging from unsupervised to supervised.~\textit{Unsupervised} corresponds to our standard \xorls~approach with the Proto exploration algorithm.~\textit{Supervised} is similar to the approach in D4RL~\citep{fu2020d4rl} and RL Unplugged~\citep{gulcehre2020rl} of using the replay buffer of a policy trained for a task-specific reward. We use TD3 as the data collection agent and refer to the reward during training as the ``data task''. We then evaluate three offline RL agents (BC, TD3+BC, and TD3) on both the data task and a different ``transfer task'' that was not seen during training. Results in the Walker environment with Stand as the data task and Walk, Run, and Flip as the transfer tasks, see~\cref{fig:unsup_vs_sup}.

For a second experiment we also add a \textit{semi-supervised} strategy as a baseline. This strategy uses a TD3 agent to maximize the sum of the extrinsic reward from the data task as well as the task-agnostic intrinsic reward from Proto.
We run this procedure on Cheetah and Walker using Run and Stand respectively as the data tasks and Run Backward and Walk as the transfer tasks, see results in~\cref{fig:supervision}.

As we hypothesized, unsupervised data allows for strong performance on both data and transfers tasks while supervised data only facilitates strong performance on the data task while failing to transfer. Semi-supervised data was capable of facilitating transfer on the easier Stand-Walk pair, but failed more clearly on the Run-Run Backward task. This is likely because Stand is a pre-requisite to Walk, while successful Run and Run Backward behaviors cover much more different parts of the state space.

Examining the offline RL algorithms, we find that TD3+BC is the strongest on the supervised data, but struggles to transfer from unsupervised data. This makes sense because the TD3+BC algorithm was designed for the D4RL benchamrk that consists of supervised-style datasets. In contrast, vanilla TD3 performs stronger on the unsupervised data.
Unsurprisingly, BC can work fairly well on supervised data, but fails on the unsupervised data and the transfer tasks.

Interestingly, vanilla TD3 outperforms TD3+BC on the semi-supervised data, suggesting that data diversity can facilitate standard TD learning without offline modifications. We will build on this insight in the next subsection.


\subsection{Combining~\xorls~with supervised data}

\textbf{Q4}: Would it be useful to mix exploratory data with task-specific data? 
\textbf{A4:} \textit{Yes, it is useful to mix even a small amount of unsupervised data to facilitate transfer}.

To support this, we consider mixing datasets rather than modifying the data collection algorithms. We consider the same pairs of data and transfer tasks as in the prior subsection, but this time we mix the supervised and unsupervised datasets themselves. To do this, we randomly sample trajectories from the supervised and unsupervised datasets with varying proportions of unsupervised data while keeping the size of the resulting mixed dataset constant at 1M transitions (1000 episodes). We can then see the performance of BC, TD3+BC, and TD3 on datasets with varying proportions of unsupervised data in~\cref{fig:mixing}.

Specifically, on the transfer tasks, TD3 achieves nearly the same performance with 25\% unsupervised data as with 100\%. Similarly, on the data task adding a relatively small amount of unsupervised data allows TD3 to match the performance of TD3+BC on the fully supervised data. This sends a very hopeful message that even adding a small amount of unsupervised exploratory data can facilitate both transfer to novel tasks and prevention of issues of extrapolation error for vanilla RL algorithms in the offline setting. 

Comparing the offline RL algorithms more directly, we see that TD3 effectively takes advantage of the mixed datasets while TD3+BC and BC do not. By attempting to constrain to be near the behavior policy, algorithms like TD3+BC and BC can struggle to deal with data that is generated by a mixture of different behaviors, especially when these behaviors are not optimizing the test-time task.

\begin{figure*}[h]
    \centering
    
    \begin{subfigure}{\linewidth}
		\centering
		\includegraphics[width=0.9\linewidth]{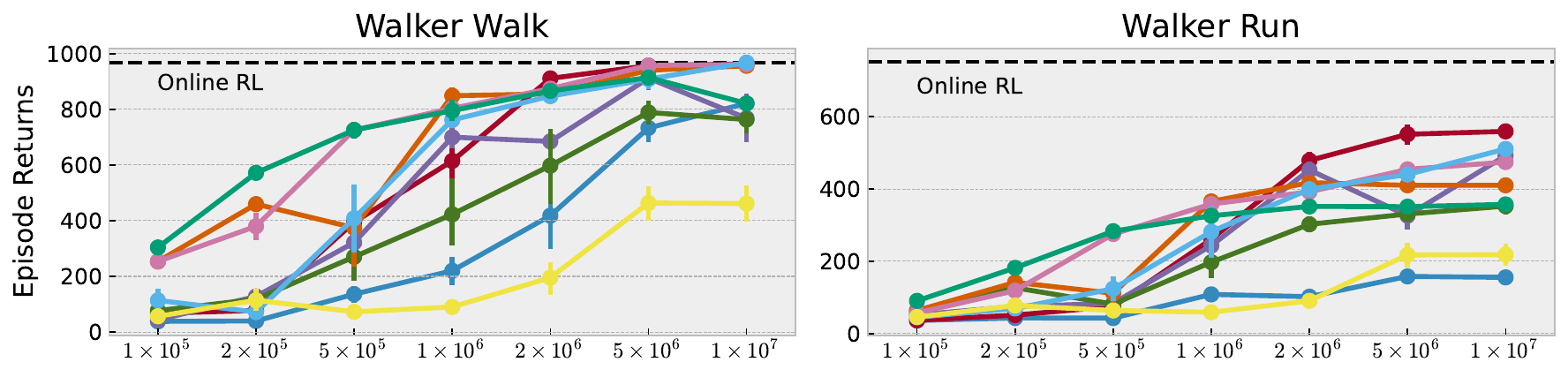}\\
    \cblock{52}{138}{189}\hspace{1mm}Random\hspace{1.5mm}
    \cblock{166}{6}{40}\hspace{1mm}ICM\hspace{1.5mm}
    \cblock{122}{104}{166}\hspace{1mm}Disagreement\hspace{1.5mm}
    \cblock{70}{120}{33}\hspace{1mm}RND\hspace{1.5mm}
    \cblock{213}{94}{0}\hspace{1mm}Proto\hspace{1.5mm}
    \cblock{204}{121}{167}\hspace{1mm}APT\hspace{1.5mm}
    \cblock{86}{180}{233}\hspace{1mm}APS\hspace{1.5mm}
    \cblock{0}{158}{115}\hspace{1mm}DIAYN\hspace{1.5mm}
    \cblock{240}{228}{66}\hspace{1mm}SMM
    \caption{Correlation between TD3 performance and the unsupervised data collection transition budget.}
        \label{fig:ds_size_data}
     \end{subfigure}
     		
   \begin{subfigure}{\linewidth}
		\centering
		\includegraphics[width=0.9\linewidth]{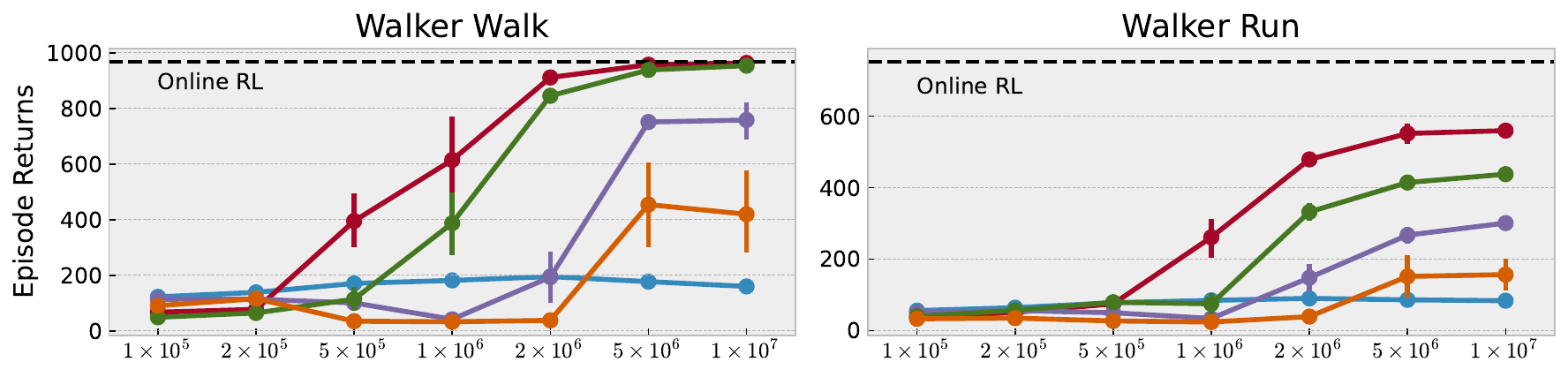}\\
		\cblock{52}{138}{189}\hspace{1mm}BC\hspace{1.5mm}
    \cblock{213}{94}{0}\hspace{1mm}TD3+BC\hspace{1.5mm}
    \cblock{122}{104}{166}\hspace{1mm}CRR\hspace{1.5mm}
    \cblock{70}{120}{33}\hspace{1mm}CQL\hspace{1.5mm}
    \cblock{166}{6}{40}\hspace{1mm}TD3\hspace{1.5mm}
		\caption{Correlation between performance of various offline RL algorithms and unsupervised data collection transition budget for ICM. }
        \label{fig:ds_size_eval}
     \end{subfigure}

	\caption{Testing the impact of the size of unsupervised data collection budget on downstream offline RL performance.~\cref{fig:ds_size_data} provides a breakdown by datasets, while~\cref{fig:ds_size_eval} provides a breakdown by offline RL algorithms. Generally, larger datasets demonstrate better performance across both exploration and offline RL algorithms.}
        \label{fig:ds_size}
  
     \vspace{-0.5cm}
\end{figure*}

\subsection{Scaling of \xorls}

\textbf{Q5}: How effective is~\xorls~as we scale the data collection budget?
\textbf{A1:} \textit{In general, all algorithms see improved performance with increased data collection budget}.

To confirm, this we collect datasets in the Walker domain with data collection budgets ranging from 100k to 10M transitions (100 to 10k episodes) using all nine unsupervised data collection algorithms. We then run offline RL on these datasets for two different rewards: Walk and Run. To compare the different datasets we run offline TD3 on each (since TD3 was the strongest performer in prior experiments). Then to compare the offline RL algorithms, we run each of our five offline RL algorithms on the datasets generated by ICM (since ICM was the strongest performing dataset with TD3). Results are in \cref{fig:ds_size}.

First, we evaluate the results across the different exploration algorithms. All algorithms see returns to scale, but we do observe gains beginning to saturate by 10M transitions on the Run task before reaching optimal performance suggesting there is still room for improved algorithms. In low-data settings we see the best performance from DIAYN while in the high-data regime ICM begins to dominate. We note that the random agent also sees return to scale, suggesting that the Walk task is not a difficult exploration problem.

We can also compare across the different offline RL algorithms on the ICM data. Again we see clear returns to scale as performance improves with dataset size. As in previous experiments we see that TD3 is the strongest algorithm and this holds true across dataset sizes.

\section{Discussion}

In this paper we propose~\xorls~-- a novel framework for exploratory data collection for offline RL. Through an extensive empirical evaluation we
conclude that using diverse data can greatly simplify offline RL algorithms by removing the need to fight the extrapolation problem. We then demonstrate that exploratory data is more suitable for efficient multi-task offline RL than standard task-directed offline RL data. Finally, we release~\xorls's datasets as well as the experimentation code to encourage future research in this area. 
Now we will discuss a few of the possible directions for future research with \xorls. 

\paragraph{Designing adaptive offline RL algorithms.} Our dataset release provides a new testbed for improving offline RL algorithms. Our results suggest that previously designed offline RL algorithms perform well on supervised data, but are beaten by TD3 on unsupervised \xorls~data. Ideally, an offline RL algorithm would be able to automatically adapt to the given dataset to recover the best of both worlds. There is a growing literature on offline hyperparameter tuning for offline RL that may be useful for approaching this challenge \citep{paine2020hyperparameter, kumar2021workflow, zhang2021towards, fu2021benchmarks, lee2021model}.

\paragraph{Improving unsupervised data collection.}
As a side benefit, our approach yields a novel evaluation protocol for unsupervised RL by evaluating the generated dataset directly via offline RL.
This differs from prior work on unsupervised RL \cite{laskin2021urlb} by emphasizing the dataset rather than the pre-trained policy. This allows for comparisons across different unsupervised RL algorithms without having to fine-tune online. In future work we hope to explore how to optimize the unsupervised exploration algorithms to improve performance of the downstream offline RL. \xorls~provides a useful starting point and framework for this problem.

\paragraph{Adding exploratory data to improve stability.}
Our preliminary experiments suggest that adding even a small amount of unsupervised data can allow vanilla TD3 to outperform other offline RL algorithms. Future work in this direction could take this insight further. For example, currently most offline RL algorithms are essentially incorporating some sort of regularization into the algorithm, but our results suggest it could be fruitful to instead ``regularize'' the data itself by mixing in some exploratory data. This is potentially a useful and practical technique that is under-explored in the literature.

\paragraph{Scaling \xorls~to more challenging environments.}
While we offer a large scale study on several domains from the DeepMind control suite, we do acknowledge that it would be fruitful to scale our framework to more complicated environments and settings. Specifically, we would be interested to extend these ideas into challenging exploration problems like manipulation, and higher-dimensional or even image-based observation spaces.


\subsection*{Acknowledgments}
This work was supported by grants from Honda, NSF AI4OPT AI Institute for Advances in Optimization under NSF 2112533, and ONR awards N00014-21-1-2758 and N00014-21-1-2769.
DB is supported by the Department of Defense (DoD) through the National Defense Science \& Engineering Graduate Fellowship (NDSEG) Program. The authors would also like to thank Mahi Shafiullah and Ben Evans for comments on a draft of the paper.

\bibliography{main}
\bibliographystyle{icml2022}

\newpage





\includeappendixtrue 

\ifincludeappendix

\onecolumn
\appendix
\section*{Appendix}

\section{Hyper-parameters}
\label{app:hyperparams}
In this section we comprehensively describe configuration of both unsupervised RL and offline RL algorithms.
\subsection{Unsupervised RL Hyper-parameters}
We adhere closely to the parameter settings from URLB~\citep{laskin2021urlb}. We list the common hyper-parameters in~\cref{table:url_common_hp} and per-algorithm hyper-parameters in~\cref{table:url_individual_hp}. 

\begin{table}[h!]
\caption{\label{table:url_common_hp} Common hyper-parameters for unsupervised RL algorithms.}
\centering
\begin{tabular}{lc}
\hline
Common hyper-parameter       & Value \\
\hline
\: Replay buffer capacity & $10^6$ \\
\: Seed frames & $4000$ \\
\: Mini-batch size & $1024$ \\
\: Discount ($\gamma$) & $0.99$ \\
\: Optimizer & Adam \\
\: Learning rate & $10^{-4}$ \\
\: Agent update frequency & $2$ \\
\: Critic target EMA rate ($\tau_Q$) & $0.01$ \\
\: Hidden dim. & $1024$ \\
\: Exploration stddev clip & $0.3$ \\
\: Exploration stddev value & $0.2$ \\

\hline
\end{tabular}

\end{table}

\begin{table}[h!]
\caption{\label{table:url_individual_hp} Individual hyper-parameters for unsupervised RL algorithms.}
\centering
\begin{tabular}{lc}
\hline
Random hyper-parameter       & Value \\
\hline
\: Policy distribution & $\mathrm{uniform}$ \\
\hline
ICM hyper-parameter       & Value \\
\hline
\: Reward transformation & $\log (r + 1.0)$ \\
\: Forward net arch. & $(|\mathcal{O}| + |\mathcal{A}|) \to 1024 \to 1024 \to |\mathcal{O}|$ $\textrm{ReLU}$ MLP\\
\: Inverse net arch. & $(2 \times |\mathcal{O}|) \to 1024 \to 1024 \to |\mathcal{A}|$ $\textrm{ReLU}$ MLP\\
\hline
Disagreement hyper-parameter       & Value \\
\hline
\: Ensemble size & $5$ \\
\: Forward net arch: & $(|\mathcal{O}| + |\mathcal{A}|) \to 1024 \to 1024 \to |\mathcal{O}|$ $\textrm{ReLU}$ MLP\\
\hline
RND hyper-parameter       & Value \\
\hline
\: Predictor \& target net arch. & $|\mathcal{O}| \to 1024 \to 1024 \to 512$ $\textrm{ReLU}$ MLP\\
\: Normalized observation clipping & 5 \\
\hline
APT hyper-parameter       & Value \\
\hline
\: Reward transformation & $\log (r + 1.0)$ \\
\: Forward net arch. & $(512 + |\mathcal{A}|) \to 1024 \to 512$ $\textrm{ReLU}$ MLP\\
\: Inverse net arch. & $(2 \times 512) \to 1024 \to |\mathcal{A}|$ $\textrm{ReLU}$ MLP\\
\: $k$ in $\mathrm{NN}$ & $12$ \\
\: Avg top $k$ in $\mathrm{NN}$ & True \\
\hline
Proto hyper-parameter       & Value \\
\hline
\: Predictor dim. & $128$ \\
\: Projector dim. & $512$ \\ 
\: Number of prototypes & $512$ \\
\: Softmax temperature & $0.1$ \\
\: $k$ in $\mathrm{NN}$ & $3$ \\
\: Number of candidates per prototype & $4$ \\
\: Encoder target EMA rate ($\tau_\mathrm{enc}$) & $0.05$ \\
\hline
SMM hyper-parameter & Value \\
\hline 
\: Skill dim. & $4$ \\
\: Skill discriminator learning rate & $10^{-3}$ \\
\: VAE lr & $10^{-2}$ \\

\hline
DIAYN hyper-parameter       & Value \\
\hline

\: Skill dim & 16 \\ 
\: Skill sampling frequency (steps) & 50 \\
\: Discriminator net arch. & $512 \to 1024 \to 1024 \to 16$ $\textrm{ReLU}$ MLP  \\
\hline 
APS hyper-parameter       & Value \\
\hline
\: Reward transformation & $\log (r + 1.0)$ \\
\: Successor feature dim. & $10$ \\
\: Successor feature net arch. & $|\mathcal{O}| \to 1024 \to 1024 \to 10$ $\textrm{ReLU}$ MLP\\
\: $k$ in $\mathrm{NN}$ & $12$ \\
\: Avg top $k$ in $\mathrm{NN}$ & True \\
\: Least square batch size & $4096$ \\
\hline

\end{tabular}

\end{table}

\clearpage

\subsection{Offline RL Hyper-parameters}
In~\cref{table:orl_common_hp} we present a common set of hyper-parameters used in our experiments, while in~\cref{table:orl_individual_hp}
we list individual hyper-parameters for each method. To select hyper-parameters for offline RL algorithms we started with from the original hyper-parameters presented in the corresponding papers. We then performed a grid search over the most important parameters (i.e. learning rate, $\alpha$, mini-batch size, etc.) on a supervised dataset collected by TD3~\citep{fujimoto2018addressing} on Cheetah Run. 
\paragraph{TD3+BC} We swept over learning rate $[10^{-3}, 3\cdot10^{-4}, 10^{-4}]$, mini-batch size $[256, 512, 1024]$ and $\alpha$ $[0.01, 0.1, 1.0, 2.5, 5.0, 10.0, 50.0]$. We also found it useful to increase the dimension of hidden layers to $1024$. 
\paragraph{CRR} We swept over the transformation function (indicator, exponent, identity), the number of samples to estimate value function $[5, 10, 20]$ as well as learning rate $[10^{-3}, 3\cdot10^{-4}, 10^{-4}]$ and mini-batch size $[256, 512, 1024]$.
\paragraph{CQL} We swept over $\alpha$ $[0.01, 0.1, 0.5, 1.0, 2.0, 5.0, 10.0]$, the number of sampled actions $[3, 5, 10, 20]$, learning rate $[10^{-3}, 3\cdot10^{-4}, 10^{-4}]$ and mini-batch size $[256, 512, 1024]$.

\begin{table}[h]
\caption{\label{table:orl_common_hp} Common hyper-parameters for offline RL algorithms.}
\centering
\begin{tabular}{lc}
\hline
Common hyper-parameter       & Value \\
\hline
\: Replay buffer capacity & $10^6$ \\
\: Mini-batch size & $1024$ \\
\: Discount ($\gamma$) & $0.99$ \\
\: Optimizer & Adam \\
\: Learning rate & $10^{-4}$ \\
\: Agent update frequency & $2$ \\
\: Training steps & $5\times 10^5$ \\
\hline
\end{tabular}

\end{table}

\begin{table}[h]
\caption{\label{table:orl_individual_hp} Individual hyper-parameters for offline RL algorithms.}
\centering
\begin{tabular}{lc}
\hline
BC hyper-parameter       & Value \\
\hline
\: Num hidden layers & $2$ \\ 
\: Hidden dim. & $1024$ \\
\hline
TD3+BC hyper-parameter       & Value \\
\hline
\: $\alpha$ &$2.5$ \\
\: Critic target EMA rate ($\tau_Q$) & $0.01$ \\
\: Num hidden layers & $2$ \\ 
\: Hidden dim. & $1024$ \\
\hline
CRR hyper-parameter       & Value \\
\hline
\: Num samples to estimate $V$ & $10$ \\ \: Critic target EMA rate ($\tau_Q$) & $0.01$ \\
\: Transformation & indicator \\ 
\: Num hidden layers & $2$ \\ 
\: Hidden dim. & $1024$ \\
\hline
CQL hyper-parameter       & Value \\
\hline
\: $\alpha$ & $0.01$ \\ 
\: Critic target EMA rate ($\tau_Q$) & $0.01$ \\
\: Lagrange & False \\ 
\: Num sampled actions & $3$ \\ 
\: Num hidden layers & $2$ \\ 
\: Hidden dim. & $1024$ \\
\hline
TD3 hyper-parameter       & Value \\
\hline
\: Stddev clip & $0.3$ \\
\: Critic target EMA rate ($\tau_Q$) & $0.01$ \\
\: Num hidden layers & $2$ \\ 
\: Hidden dim. & $1024$ \\
\hline

\end{tabular}

\end{table}
\clearpage
\newpage

\section{Full Results on the PointMass Experiment}
\label{section:app_pm_full_results}

\begin{figure*}[h]

		\centering
		\includegraphics[width=\linewidth]{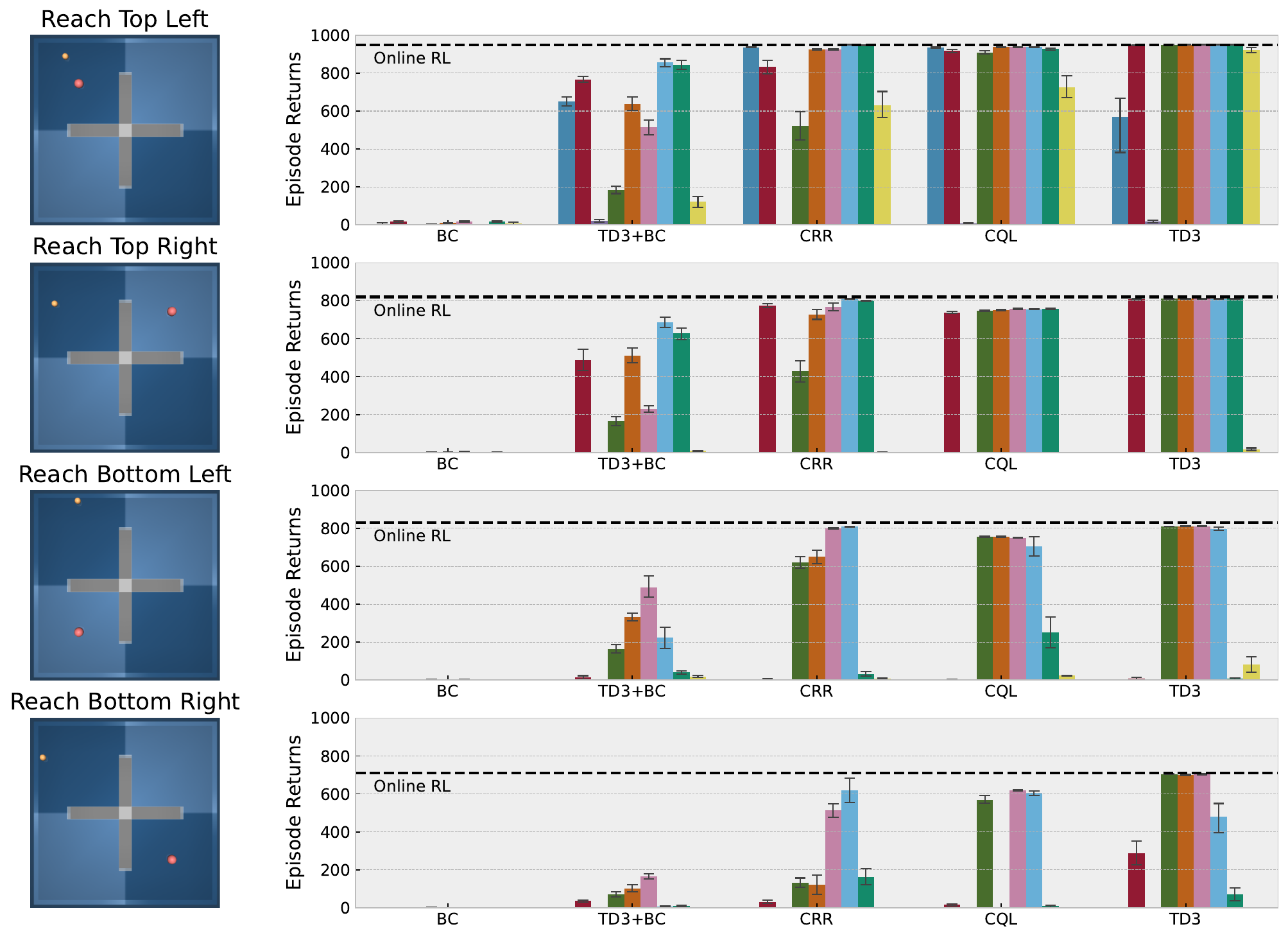}
		\hspace{1.5mm}
    \cblock{52}{138}{189}\hspace{1mm}Random\hspace{1.5mm}
    \cblock{166}{6}{40}\hspace{1mm}ICM\hspace{1.5mm}
    \cblock{122}{104}{166}\hspace{1mm}Disagreement\hspace{1.5mm}
    \cblock{70}{120}{33}\hspace{1mm}RND\hspace{1.5mm}
    \cblock{213}{94}{0}\hspace{1mm}Proto\hspace{1.5mm}
    \cblock{204}{121}{167}\hspace{1mm}APT\hspace{1.5mm}
    \cblock{86}{180}{233}\hspace{1mm}APS\hspace{1.5mm}
    \cblock{0}{158}{115}\hspace{1mm}DIAYN\hspace{1.5mm}
    \cblock{240}{228}{66}\hspace{1mm}SMM

        \label{fig:pm_eval_full}

    \caption{ Extended figures from~\cref{fig:pm}. }

\end{figure*}

\newpage

\section{Full Results on Singletask Settings}
\label{section:mult_env_full}

\begin{figure*}[h]

		\centering
		\includegraphics[width=\linewidth]{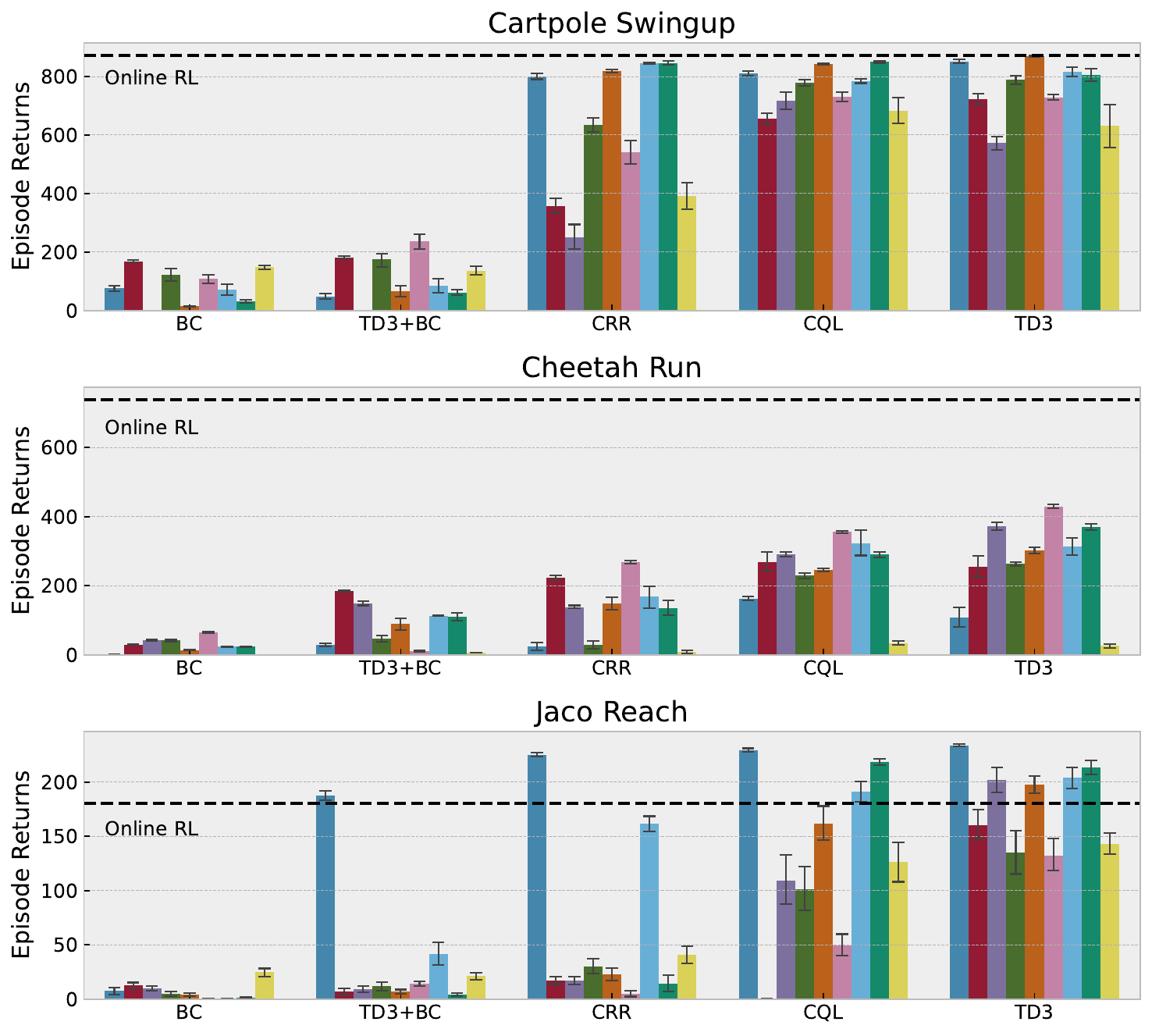}
		\hspace{1.5mm}
    \cblock{52}{138}{189}\hspace{1mm}Random\hspace{1.5mm}
    \cblock{166}{6}{40}\hspace{1mm}ICM\hspace{1.5mm}
    \cblock{122}{104}{166}\hspace{1mm}Disagreement\hspace{1.5mm}
    \cblock{70}{120}{33}\hspace{1mm}RND\hspace{1.5mm}
    \cblock{213}{94}{0}\hspace{1mm}Proto\hspace{1.5mm}
    \cblock{204}{121}{167}\hspace{1mm}APT\hspace{1.5mm}
    \cblock{86}{180}{233}\hspace{1mm}APS\hspace{1.5mm}
    \cblock{0}{158}{115}\hspace{1mm}DIAYN\hspace{1.5mm}
    \cblock{240}{228}{66}\hspace{1mm}SMM

        \label{fig:mult_env_full}

    \caption{Offline evaluation of unsupervised datasets on one task for each of three different domains. Here we choose four representative unsupervised exploration algorithms. }

\end{figure*}

\newpage

\section{Full Results on Multitask Settings}
\label{section:mult_task_full}

\begin{figure*}[h]

		\centering
		\includegraphics[width=\linewidth]{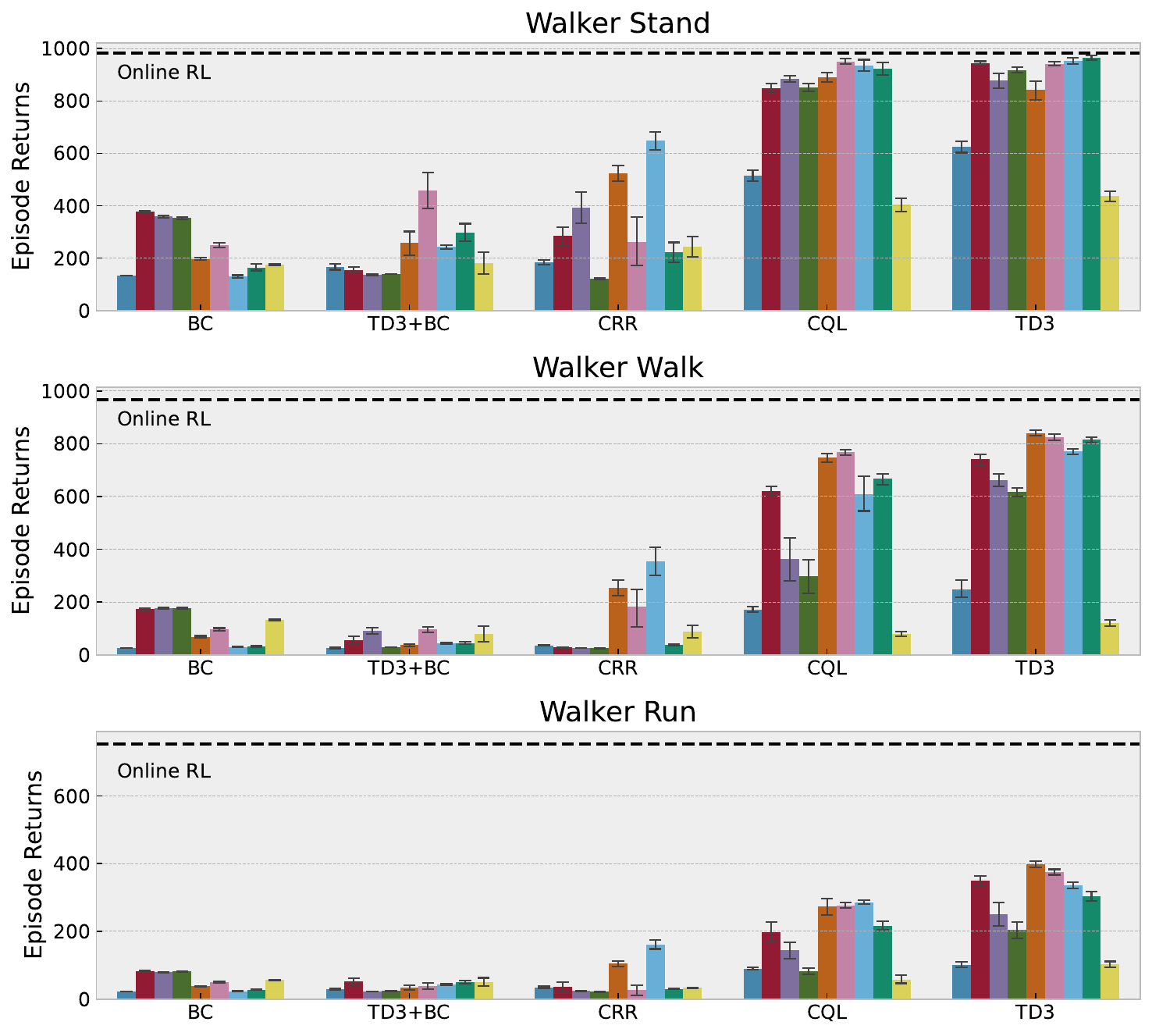}
		\hspace{1.5mm}
    \cblock{52}{138}{189}\hspace{1mm}Random\hspace{1.5mm}
    \cblock{166}{6}{40}\hspace{1mm}ICM\hspace{1.5mm}
    \cblock{122}{104}{166}\hspace{1mm}Disagreement\hspace{1.5mm}
    \cblock{70}{120}{33}\hspace{1mm}RND\hspace{1.5mm}
    \cblock{213}{94}{0}\hspace{1mm}Proto\hspace{1.5mm}
    \cblock{204}{121}{167}\hspace{1mm}APT\hspace{1.5mm}
    \cblock{86}{180}{233}\hspace{1mm}APS\hspace{1.5mm}
    \cblock{0}{158}{115}\hspace{1mm}DIAYN\hspace{1.5mm}
    \cblock{240}{228}{66}\hspace{1mm}SMM

        \label{fig:mult_task_full}

    \caption{ Offline evaluation of datasets for the Walker environment under three different rewards (Stand, Walk, and Run). We observe that ExORL allows for data relabeling to enable multi-task offline RL. }

\end{figure*}

\newpage

\section{Offline RL on Suffixes of Unsupervised Datasets}
\label{section:suffix}
In~\cref{fig:suffix} we demonstrate that it is important to use the entire replay buffers collected by~\xorls, rather than only training on the later parts of this data.

\begin{figure*}[h]
    \centering

    \includegraphics[width=0.9\linewidth]{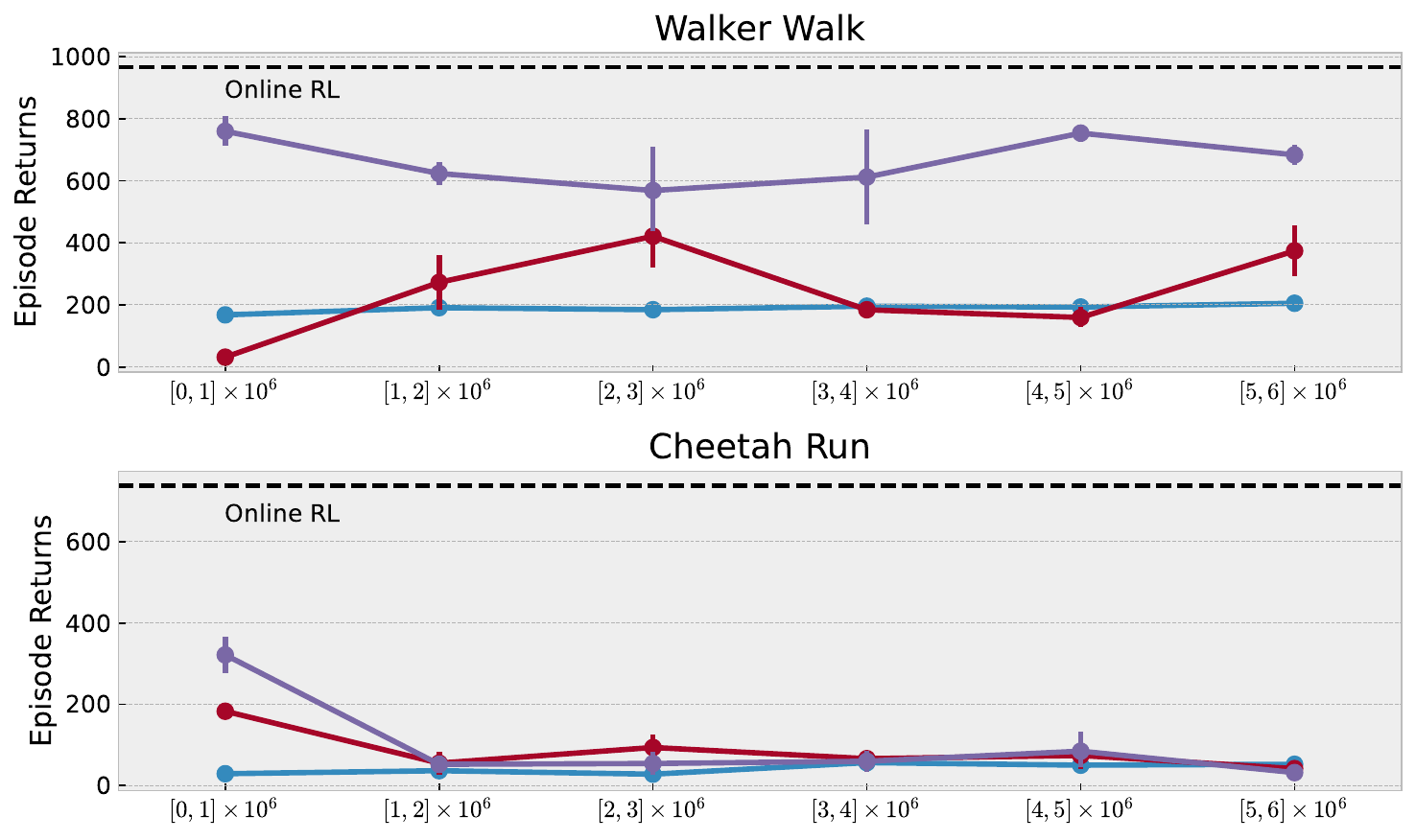}\\
        \hspace{1.5mm} \cblock{52}{138}{189}\hspace{1mm}BC\hspace{1.5mm}
    \cblock{166}{6}{40}\hspace{1mm}TD3+BC\hspace{1.5mm}
    \cblock{122}{104}{166}\hspace{1mm}TD3\hspace{1.5mm}
    
    \caption{Instad of training on the full replay buffer collected by ICM, we perform offline RL training on various suffixes of the replay buffer. The x-axis represents segments of the replay buffuer. We observe that it is important to keep around early transitions.}
    \label{fig:suffix}
\end{figure*}

\newpage
\section{Compute Resources}
\label{app:compute}

\xorls~is designed to be accessible to the RL research community. We provide an efficient implementation of our framework, including data collection, relabeling, and offline RL evaluation, that requires a single GPU. For local debugging experiments we used NVIDIA RTX GPUs. For large-scale runs used to generate all results in this manuscripts, we used NVIDIA Tesla V100 GPU instances. All experiments were run on internal clusters. Each offline RL algorithm trains in roughly 2 hours for 500k gradient steps on the collected datasets.

\section{The ExORL Environments and Tasks}
\label{app:dmc_tasks}
We provide a summary of used environments and tasks in our paper in~\cref{table:benchamrks}.

\begin{table}[!h]
\centering
\begin{tabular}{lccccc}
\hline
Environment & Task & Traits  &$\mathrm{dim}(\mathcal{S})$ & $\mathrm{dim}(\mathcal{A})$   \\

\hline
Walker & Stand & dense reward, easy exploration, locomotion   & $18$ & $6$  \\
 & Walk & dense reward, medium exploration, locomotion   & $18$ & $6$  \\
  & Run & dense reward, hard exploration locomotion   & $18$ & $6$  \\
   & Flip & dense reward, medium exploration, locomotion   & $18$ & $6$  \\

\hline

Cheetah & Run & dense reward, hard exploration, locomotion   & $24$ & $6$  \\
 & Run Backward & dense reward, hard exploration, hard transfer   & $24$ & $6$  \\
\hline

\hline
PointMass Maze & Reach Top Left & sparse reward, hard exploration, hard transfer   & $4$ & $2$  \\

 & Reach Top Right & sparse reward, hard exploration, hard transfer    & $4$ & $2$  \\
  & Reach Bottom Left & sparse reward, hard exploration, hard transfer    & $4$ & $2$  \\
   & Reach Bottom Right & sparse reward, hard exploration, hard transfer    & $4$ & $2$  \\

\hline

Jaco & Reach & sparse reward, hard exploration, manipulation   & $55$ & $6$  \\
\hline

 Cartpole & Swingup  & dense reward, easy exploration   & $4$ & $1$  \\
\hline

\hline
\end{tabular}
\caption{\label{table:benchamrks} A detailed description of used environments and tasks from the DeepMind control suite~\citep{tassa2018deepmind}.}
\end{table}

\fi

\end{document}